\documentclass{article}
\usepackage{graphics,graphicx,caption,float,subcaption,booktabs,xcolor,multirow,array,color,ifthen,tabu,colortbl,dblfloatfix,url,xparse,mathtools,patchcmd}
\usepackage{algorithm, algorithmic,amssymb,xspace,nicefrac,microtype}
\usepackage{amsmath,amsfonts,bm}
\usepackage{ragged2e,tikz,stackengine,etoolbox,xpatch,enumerate,xstring,setspace,tabularx,makecell,changepage,cuted,titlesec}
\usepackage{natbib}
\usepackage[pagebackref=true,breaklinks=true,colorlinks=true,bookmarks=false]{hyperref}
\usepackage[capitalise,noabbrev,nameinlink]{cleveref}
\usepackage[english]{babel}  %
\usepackage[accepted]{icml2020}
\usepackage{enumitem}  %

\titlespacing*{\paragraph}{0pt}{0ex plus .1ex}{1em}

\DeclareMathOperator*{\argmax}{arg\,max}

\DeclarePairedDelimiterX{\RoundBrackets}[1]{(}{)}{#1}
\NewDocumentCommand{\pr}{ O{p} r() }{\def\prArg{#2}\patchcmd{\prArg}{|}{\mid}{}{}#1\RoundBrackets{\prArg}}
\NewDocumentCommand{\p}{ r() }{\pr[p](#1)}
\NewDocumentCommand{\q}{ r() }{\pr[q](#1)}
\NewDocumentCommand{\pol}{ r() }{\ensuremath{\pr[\pi](#1)}}
\NewDocumentCommand{\val}{ r() }{\ensuremath{V(#1)}}
\NewDocumentCommand{\Normal}{ r() }{\pr[\operatorname{\mathcal{N}}](#1)}
\NewDocumentCommand{\Cat}{ r() }{\pr[\operatorname{Cat}](#1)}
\NewDocumentCommand{\Bin}{ r() }{\pr[\operatorname{Bin}](#1)}
\NewDocumentCommand{\Beta}{ r() }{\pr[\operatorname{Beta}](#1)}
\NewDocumentCommand{\Bernoulli}{ r() }{\pr[\operatorname{Bernoulli}](#1)}
\NewDocumentCommand{\Dir}{ r() }{\pr[\operatorname{Dir}](#1)}

\NewDocumentCommand{\I}{ r() }{\pr[\operatorname{I}](#1)}
\RenewDocumentCommand{\H}{ r() }{\pr[\operatorname{H}](#1)}

\newlength\widthE

\creflabelformat{equation}{#2\textup{#1}#3}
\crefname{algocf}{Algorithm}{Algorithms}
\Crefname{algocf}{Algorithm}{Algorithms}

\icmltitlerunning{Planning to Explore via Self-Supervised World Models}

\begin{document}

\twocolumn[
\icmltitle{Planning to Explore via Self-Supervised World Models}
\icmlsetsymbol{equal}{*}
\begin{icmlauthorlist}
\icmlauthor{Ramanan Sekar}{penn,equal}
\icmlauthor{Oleh Rybkin}{penn,equal}
\icmlauthor{Kostas Daniilidis}{penn}
\icmlauthor{Pieter Abbeel}{ucb}
\icmlauthor{Danijar Hafner}{google,uoft}
\icmlauthor{Deepak Pathak}{cmu,fair}
\end{icmlauthorlist}
\icmlaffiliation{ucb}{UC Berkeley}
\icmlaffiliation{cmu}{Carnegie Mellon University}
\icmlaffiliation{penn}{University of Pennsylvania}
\icmlaffiliation{fair}{Facebook AI Research}
\icmlaffiliation{google}{Google Research, Brain Team}
\icmlaffiliation{uoft}{University of Toronto}
\icmlcorrespondingauthor{Oleh Rybkin}{oleh@seas.upenn.edu}
\icmlkeywords{}
\vskip 0.3in
]
\printAffiliationsAndNotice{\icmlEqualContribution}

\begin{abstract}
\begin{hyphenrules}{nohyphenation}
Reinforcement learning allows solving complex tasks, however, the learning tends to be task-specific and the sample efficiency remains a challenge. We present Plan2Explore, a self-supervised reinforcement learning agent that tackles both these challenges through a new approach to self-supervised exploration and fast adaptation to new tasks, which need not be known during exploration. During exploration, unlike prior methods which retrospectively compute the novelty of observations after the agent has already reached them, our agent acts efficiently by leveraging planning to seek out expected future novelty. After exploration, the agent quickly adapts to multiple downstream tasks in a zero or a few-shot manner. We evaluate on challenging control tasks from high-dimensional image inputs. Without any training supervision or task-specific interaction, Plan2Explore outperforms prior self-supervised exploration methods, and in fact, almost matches the performances oracle which has access to rewards. Videos and code: \url{https://ramanans1.github.io/plan2explore/}
\end{hyphenrules}
\end{abstract}

\section{Introduction}
The dominant approach in sensorimotor control is to train the agent on one or more pre-specified tasks either via rewards in reinforcement learning, or via demonstrations in imitation learning. However, learning each task from scratch is often inefficient, requiring a large amount of task-specific environment interaction for solving each task. How can an agent quickly generalize to unseen tasks it has never experienced before in a zero or few-shot manner?

\begin{figure}[t!]
\centering
\includegraphics[width=\linewidth]{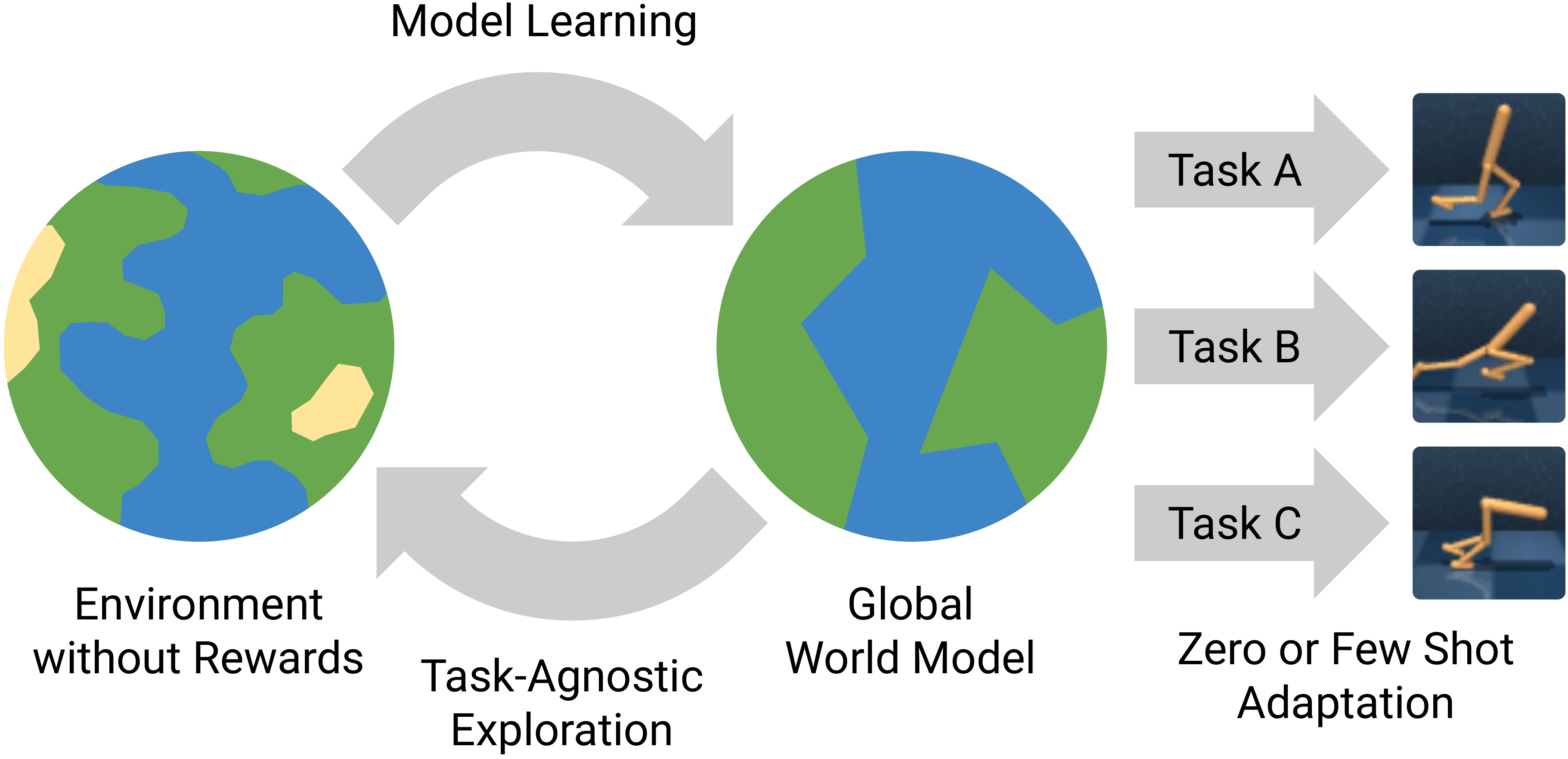}
\vspace{-6mm}
\caption{The agent first leverages planning to explore in a self-supervised manner, without task-specific rewards, to efficiently learn a global world model. After the exploration phase, it receives reward functions at test time to adapt to multiple downstream tasks, such as standing, walking, running, and flipping using either zero or few tasks-specific interactions.}
\label{fig:teaser}
\end{figure}

\vspace{-0.08in}
\paragraph{Task-agnostic RL}
Because data collection is often expensive, it would be ideal to not keep collecting data for each new task. Hence, we explore the environment once without reward to collect a diverse dataset for later solving any downstream task, as shown in \cref{fig:teaser}. After the task-agnostic exploration phase, the agent is provided with downstream reward functions and needs to solve the tasks with limited or no further environment interaction. Such a self-supervised approach would allow solving various tasks without having to repeat the expensive data collection for each new task.

\vspace{-0.08in}
\paragraph{Intrinsic motivation}
To explore complex environments in the absence of rewards, the agent needs to follow a form of intrinsic motivation that is computed from inputs that could be high-dimensional images. For example, an agent could seek inputs that it cannot yet predict accurately~\cite{schmidhuber1991possibility,oudeyer2007intrinsic,pathakICMl17curiosity}, maximally influence its inputs~\cite{klyubin2005empowerment,eysenbach2018diversity}, or visit rare states~\cite{poupart2006analytic,lehman2011evolving,bellemare2016unifying,burda2018rnd}.
However, most prior methods learn a model-free exploration policy to act in the environment
which needs large amounts of samples for finetuning or adaptation when presented with rewards for downstream tasks.

\begin{figure*}[t!]
\centering
\includegraphics[width=\textwidth]{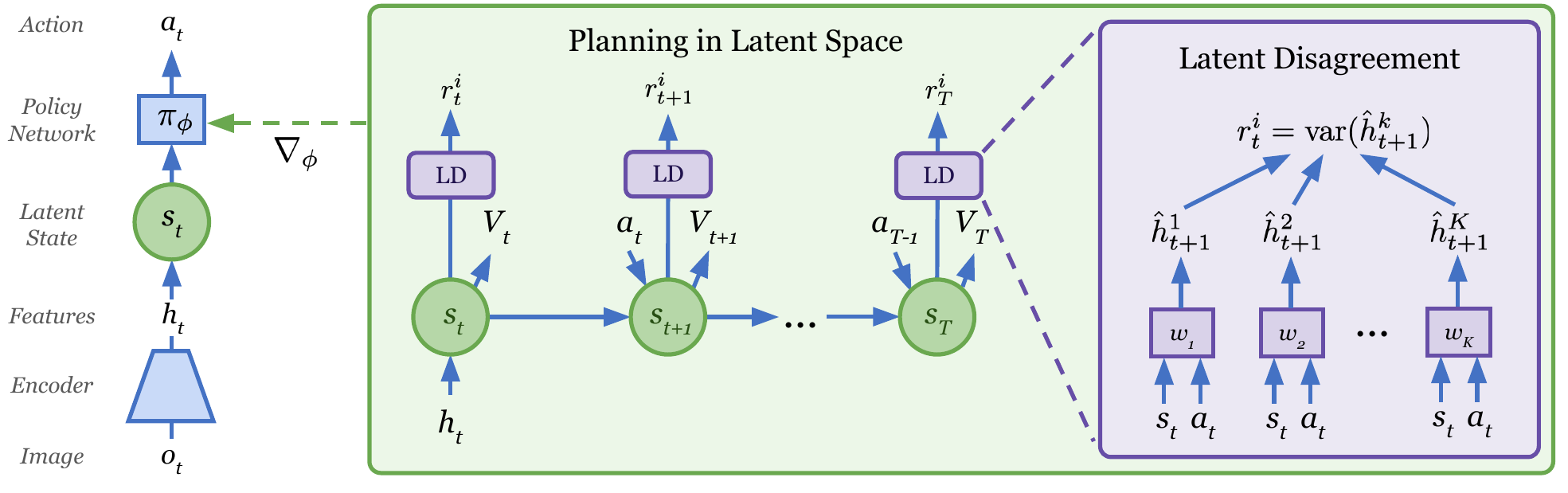}
\vspace{-9mm}
\caption{Overview of Plan2Explore. Each observation $o_t$ at time $t$ is first encoded into features $h_t$ which are then used to infer a recurrent latent state $s_t$. At each training step, the agent leverages planning to explore by imagining the consequences of the actions of policy $\pi_\phi$ using the current world model. The planning objective is to maximize expected novelty $r_t^i$ over all future time steps, computed as the disagreement in the predicted next image embedding $h_{t+1}$ from an ensemble of learned transition dynamics $w_k$. This planning objective is backpropagated all the way through the imagined rollout states to improve the exploration policy $\pi_\phi$. The learned model is used for planning to explore in latent space, and the data collected during exploration is in turn used to improve the model. This world model is then later used to plan for novel tasks at test time by replacing novelty reward with task reward.}
\vspace{-1mm}
\label{fig:method}
\end{figure*}

\vspace{-0.08in}
\paragraph{Retrospective novelty}
Model-free exploration methods not only require large amounts of experience to adapt to downstream tasks, they can also be inefficient during exploration. These agents usually first act in the environment, collect trajectories and then calculate an intrinsic reward as the agent's current estimate of novelty. This approach misses out on efficiency by operating retrospectively, that is, the novelty of inputs is computed after the agent has already reached them. For instance, in curiosity~\cite{pathakICMl17curiosity}, novelty is measured by computing error between the prediction of the next state and the ground truth only after the agent has visited the next state. Hence, it seeks out previously novel inputs that have already been visited and would not be novel anymore. Instead, one should directly seek out future inputs that are expected to be novel.

\vspace{-0.09in}
\paragraph{Planning to explore}
We address both of these challenges -- quick adaptation and expected future novelty -- within a common framework while learning directly from high-dimensional image inputs. Instead of maximizing intrinsic rewards in retrospect, we learn a world model to plan ahead and seek out the expected novelty of future situations. This lets us learn the exploration policy purely from imagined model states, without causing additional environment interaction~\cite{sun2011planning,shyam2019max}. The exploration policy is optimized purely from trajectories imagined under the model to maximize the intrinsic rewards computed by the model itself. After the exploration phase, the learned world model is used to train downstream task policies in imagination via offline reinforcement learning, without any further environment interaction.

\vspace{-0.09in}
\paragraph{Challenges}
The key challenges for planning to explore are to train an accurate world model from high-dimensional inputs and to define an effective exploration objective. We focus on world models that predict ahead in a compact latent space and have recently been shown to solve challenging control tasks from images~\cite{hafner2018planet,zhang2018solar}. Predicting future compact representations facilitates accurate long-term predictions and lets us efficiently predict thousands of future sequences in parallel for policy learning.

An ideal exploration objective should seek out inputs that the agent can learn the most from (epistemic uncertainty) while being robust to stochastic parts of the environment that cannot be learned accurately (aleatoric uncertainty). This is formalized in the expected information gain~\cite{lindley1956expinfogain}, which we approximate as the disagreement in predictions of an ensemble of one-step models. These one-step models are trained alongside the world model and mimic its transition function. The disagreement is positive for novel states, but given enough samples, it eventually reduces to zero even for stochastic environments because all one-step predictions converge to the mean of the next input~\cite{pathak2019self}.

\vspace{-0.09in}
\paragraph{Contributions}
We introduce Plan2Explore, a self-supervised reinforcement learning agent that leverages planning to efficiently explore visual environments without rewards. Across 20 challenging control tasks without access to proprioceptive states or rewards, Plan2Explore achieves state-of-the-art zero-shot and adaptation performance. Moreover, we empirically study the questions:
\begin{itemize}[noitemsep,topsep=0pt]
\item How does planning to explore via latent disagreement compare to a supervised oracle and other model-free and model-based intrinsic reward objectives?
\item How much task-specific experience is enough to finetune a self-supervised model to reach the performance of a task-specific agent?
\item To what degree does a self-supervised model generalize to unseen tasks compared to a task-specific model trained on a different task in the same environment?
\item What is the advantage of maximizing expected future novelty in comparison to retrospective novelty?
\end{itemize}

\section{Control with Latent Dynamics Models}
\label{sec:background}
World models summarize past experience into a representation of the environment that enables predicting imagined future sequences~\cite{sutton1991dyna,watter2015embed,ha2018world}. When sensory inputs are high-dimensional observations, predicting compact latent states $s_t$ lets us predict many future sequences in parallel due to memory efficiency.\footnote{The latent model states $s_t$ are not to be confused with the unknown true environment states.} Specifically, we use the latent dynamics model of PlaNet~\cite{hafner2018planet}, that consists of the following key components that are illustrated in \cref{fig:method},
\begin{gather}
\begin{aligned}
&\text{Image encoder:} &&h_t=e_\theta(o_t) \\
&\text{Posterior dynamics:} &&\pr[q_\theta](s_t|s_{t-1},a_{t-1},h_t) \\
&\text{Prior dynamics:} &&\pr[p_\theta](s_t|s_{t-1},a_{t-1}) \\
&\text{Reward predictor:} &&\pr[p_\theta](r_t|s_t) \\
&\text{Image decoder:} &&\pr[p_\theta](o_t|s_t).
\label{eq:world_model}
\end{aligned}
\end{gather}%
The image encoder is implemented as a CNN, and the posterior and prior dynamics share an RSSM~\cite{hafner2018planet}. The temporal prior predicts forward without access to the corresponding image. The reward predictor and image decoder provide a rich learning signal to the dynamics. The distributions are parameterized as diagonal Gaussians. All model components are trained jointly similar to a variational autoencoder (VAE)~\cite{kingma2013vae,rezende2014vae} by maximizing the evidence lower bound (ELBO).

Given this learned world model, we need to derive behaviors from it. Instead of online planning, we use Dreamer~\cite{hafner2019dreamer} to efficiently learn a parametric policy inside the world model that considers long-term rewards. Specifically, we learn two neural networks that operate on latent states of the model. The state-value estimates the sum of future rewards and the actor tries to maximize these predicted values,
\begin{gather}
\begin{aligned}
\text{Actor:} \quad \pol(a_t|s_t) \qquad
\text{Value:} \quad \val(s_t).
\label{eq:actor_critic}
\end{aligned}
\end{gather}%
The learned world model is used to predict the sequences of future latent states under the current actor starting from the latent states obtained by encoding images from the replay buffer. The value function is computed at each latent state and the actor policy is trained to maximize the predicted values by propagating their gradients through the neural network dynamics model as shown in \cref{fig:method}.

\section{Planning to Explore}
We consider a learning setup with two phases, as illustrated in \cref{fig:teaser}. During self-supervised exploration, the agent gathers information about the environment and summarizes this past experience in the form of a parametric world model. After exploration, the agent is given a downstream task in the form of a reward function that it should adapt to with no or limited additional environment interaction.

During exploration, the agent begins by learning a global world model using data collected so far, and then this model is in turn used to direct agent's exploration to collect more data, as described in \cref{alg:exploration}. This is achieved by training an exploration policy inside of the world model to seek out novel states. Novelty is estimated by ensemble disagreement in latent predictions made by 1-step transition models trained alongside the global recurrent world model. More details to follow in \cref{sec:disagreement}.

During adaptation, we can efficiently optimize a task policy by imagination inside of the world model, as shown in \cref{alg:adaptation}. Since our self-supervised model is trained without being biased toward a specific task, a single trained model can be used to solve multiple downstream tasks.

\subsection{Latent Disagreement}
\label{sec:disagreement}

\begin{algorithm}[t]
\caption{Planning to Explore via Latent Disagreement}
\label{alg:exploration}
\begin{algorithmic}[1]
\STATE \makebox[4.5em][l]{\textbf{initialize:}} Dataset D from a few random episodes.
\STATE \makebox[4.5em][l]{} World model M.
\STATE \makebox[4.5em][l]{} Latent disagreement ensemble E.
\STATE \makebox[4.5em][l]{} Exploration actor-critic $\pi_\mathrm{LD}$.
\WHILE{exploring}
\STATE Train M on D.
\STATE Train E on D.
\STATE Train $\pi_\mathrm{LD}$ on LD reward in imagination of M.
\STATE Execute $\pi_\mathrm{LD}$ in the environment to expand D.
\ENDWHILE
\STATE \textbf{return} Task-agnostic D and M.
\end{algorithmic}
\end{algorithm}

\begin{algorithm}[t]
\caption{Zero and Few-Shot Task Adaptation}
\label{alg:adaptation}
\begin{algorithmic}[1]
\STATE \makebox[4.5em][l]{\textbf{input:}} World model M.
\STATE \makebox[4.5em][l]{} Dataset D without rewards.
\STATE \makebox[4.5em][l]{} Reward function R.
\STATE \makebox[4.5em][l]{\textbf{initialize:}} Latent-space reward predictor \^{R}.
\STATE \makebox[4.5em][l]{} Task actor-critic $\pi_\mathrm{R}$.
\WHILE{adapting}
\STATE Distill R into \^{R} for sequences in D.
\STATE Train $\pi_\mathrm{R}$ on \^{R} in imagination of M.
\STATE Execute $\pi_\mathrm{R}$ for the task and report performance.
\STATE Optionally, add task-specific episode to D and repeat.
\ENDWHILE
\STATE \textbf{return} Task actor-critic $\pi_\mathrm{R}$.
\vspace{.3ex}
\end{algorithmic}
\end{algorithm}

To efficiently learn a world model of an unknown environment, a successful strategy should explore the environment such as to collect new experience that improves the model the most. For this, we quantify the model's uncertainty about its predictions for different latent states. An exploration policy then seeks out states with high uncertainty. The model is then trained on the newly acquired trajectories and reduces its uncertainty in these and the process is repeated.

Quantifying uncertainty is a long-standing open challenge in deep learning~\cite{mackay1992infogain,gal2016uncertainty}. In this paper, we use ensemble disagreement as an empirically successful method for quantifying uncertainty~\cite{lakshminarayanan2017simple,osband2018randomized}. As shown in \cref{fig:method}, we train a bootstrap ensemble~\cite{breiman1996bagging} to predict, from each model state, the next encoder features. The variance of the ensemble serves as an estimate of uncertainty.

Intuitively, because the ensemble models have different initialization and observe data in a different order, their predictions differ for unseen inputs. Once the data is added to the training set, however, the models will converge towards more similar predictions, and the disagreement decreases. Eventually, once the whole environment is explored, the models should converge to identical predictions.

Formally, we define a bootstrap ensemble of one-step predictive models with parameters $\{w_k \mid k \in [1;K]\}$. Each of these models takes a model state $s_t$ and action $a_t$ as input and predicts the next image embedding $h_{t+1}$. The models are trained with the mean squared error, which is equivalent to Gaussian log-likelihood,
\begin{gather}
\begin{gathered}
\text{Ensemble predictors:} \quad \q(h_{t+1}|w_k,s_t,a_t) \\[1ex]
\q(h_{t+1}|w_k,s_t,a_t) \triangleq \Normal(\mu(w_k,s_t,a_t), 1).
\end{gathered}
\label{eq:ensemble}
\end{gather}%
We quantify model uncertainty as the variance over predicted means of the different ensemble members and use this disagreement as the intrinsic reward $\mathrm{ir}_t \triangleq \operatorname{D}(s_t,a_t)$ to train the exploration policy,
\begin{gather}
\begin{aligned}
\operatorname{D}(s_t,a_t) &\triangleq \operatorname{Var}\big( \{\mu(w_k,s_t,a_t) \mid k \in [1;K] \}\big) \\
&= \frac{1}{K-1}\sum_k\big( \mu(w_k,s_t,a_t)-\mu'\big)^2, \\
\mu' &\triangleq \frac{1}{K}\sum_k\mu(w_k,s_t,a_t).
\end{aligned}
\end{gather}%
The intrinsic reward is non-stationary because the world model and the ensemble predictors change throughout exploration. Indeed, once certain states are visited by the agent and the model gets trained on them, these states will become less interesting for the agent and the intrinsic reward for visiting them will decrease.

We learn the exploration policy using Dreamer (\cref{sec:background}). Since the intrinsic reward is computed in the compact representation space of the latent dynamics model, we can optimize the learned actor and value from imagined latent trajectories without generating images. This lets us efficiently optimize the intrinsic reward without additional environment interaction. Furthermore, the ensemble of lightweight 1-step models adds little computational overhead as they are trained together efficiently in parallel across all time steps.

\subsection{Expected Information Gain}

\begin{figure*}
\centering
\includegraphics[ width=\linewidth]{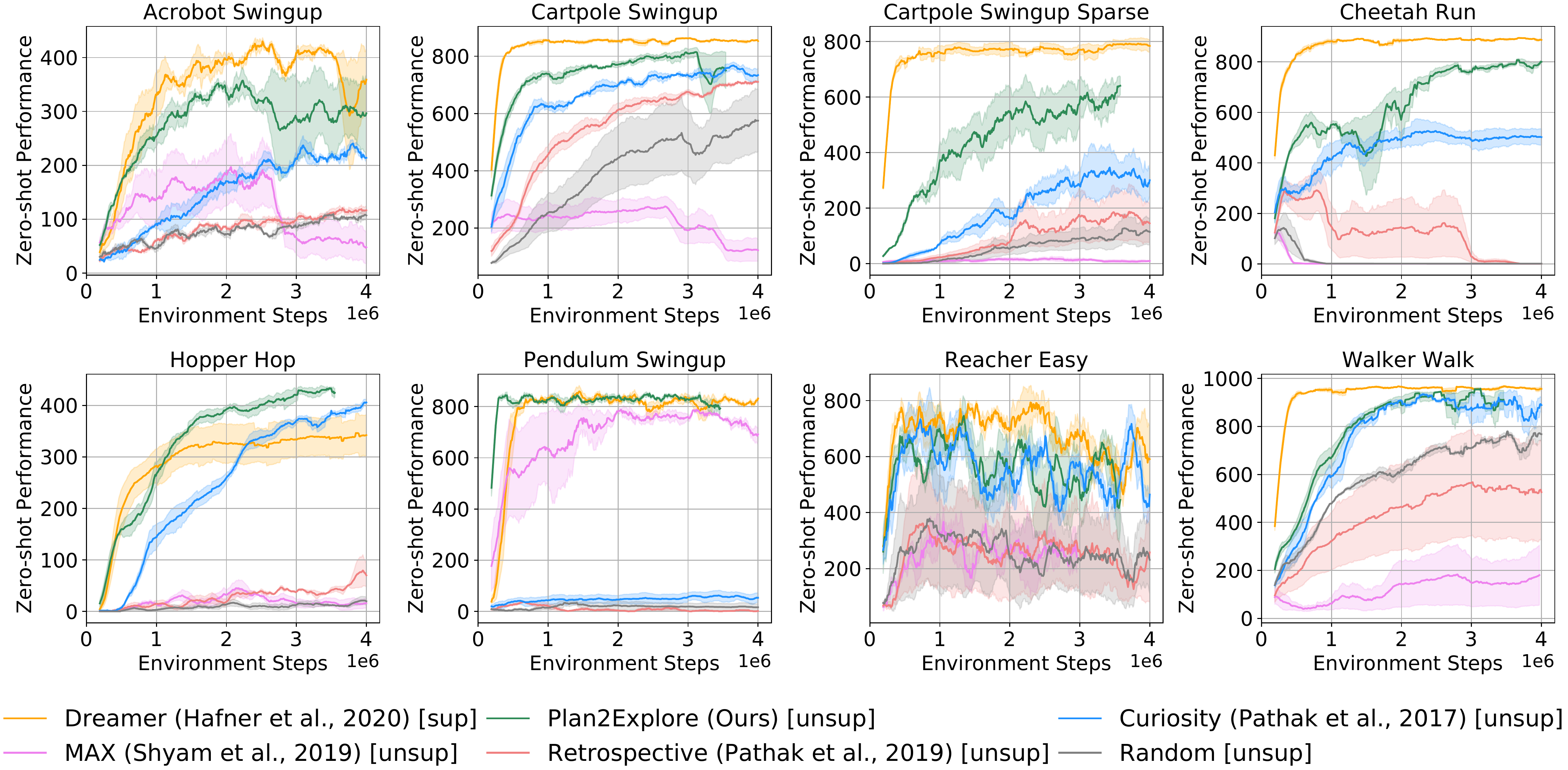}
\vspace{-0.3in}
\caption{Zero-shot RL performance from raw pixels. After training the agent without rewards, we provide it with a task by specifying the reward function at test time. Throughout the exploration, we take snapshots of the agent to train a task policy on the final task and plot its zero-shot performance. We see that Plan2Explore achieves state-of-the-art zero-shot task performance on a range of tasks, and even demonstrates competitive performance to Dreamer~\cite{hafner2019dreamer}, a state-of-the-art supervised RL agent. This indicates that Plan2Explore is able to explore and learn a global model of the environment that is useful for adapting to new tasks, demonstrating the potential of self-supervised RL. Results on all 20 tasks are in the appendix \cref{fig:zeroshot_supp} and videos on the project website.}
\label{fig:zero-shot}
\end{figure*}

Latent disagreement has an information-theoretic interpretation. This subsection derives our method from the amount of information gained by interacting with the environment, which has its roots in optimal Bayesian experiment design~\cite{lindley1956expinfogain,mackay1992infogain}.

Because the true dynamics are unknown, the agent treats the optimal dynamics parameters as a random variable $w$. To explore the environment as efficiently as possible, the agent should seek out future states that are informative of our belief over the parameters.

Mutual information formalizes the amount of bits that a future trajectory provides about the optimal model parameters on average. We aim to find a policy that shapes the distribution over future states to maximize the mutual information between the image embeddings $h_{1:T}$ and parameters $w$,
\begin{gather}
 \I(h_{t+1};w|s_t,a_t)
\end{gather}%
We operate on latent image embeddings to save computation. To select the most promising data during exploration, the agent maximizes the expected information gain,
\begin{gather}
a^*_t \triangleq \argmax_{a_t} \I(h_{t+1};w|s_t,a_t).
\end{gather}%
This expected information gain can be rewritten as conditional entropy of trajectories subtracted from marginal entropy of trajectories, which correspond to, respectively, the aleatoric and the total uncertainty of the model,
\begin{gather}
\begin{aligned}
&\I(h_{t+1};w|s_t,a_t) \\
&\quad=\H(h_{t+1}|s_t,a_t)-\H(h_{t+1}|w,s_t,a_t).
\end{aligned}
\end{gather}%
We see that the information gain corresponds to the epistemic uncertainty, i.e. the reducible uncertainty of the model that is left after subtracting the expected amount of data noise from the total uncertainty.

Trained via squared error, our ensemble members are conditional Gaussians with means produced by neural networks and fixed variances. The ensemble can be seen as a mixture distribution of parameter point masses,
\begin{gather}
\begin{aligned}
\p(w) &\triangleq \frac{1}{K} \sum_k \delta(w-w_k) \\
\p(h_{t+1}|w_k,s_t,a_t) &\triangleq \Normal(h_{t+1}|\mu(w_k,s_t,a_t),\sigma^2).
\end{aligned}\raisetag{7ex}
\end{gather}
Because the variance is fixed, the conditional entropy does not depend on the state or action in our case ($D$ is the dimensionality of the predicted embedding),
\begin{gather}
\begin{aligned}
\H(h_{t+1}|w,s_t,a_t)
&=\frac{1}{K}\sum_k \H(h_{t+1}|w_k,s_t,a_t) \\
&=\frac{D}{K}\sum_k \ln\sigma_k(s_t,a_t) + \text{const}.
\end{aligned}
\end{gather}%
We note that this fixed variance approach is applicable even in environments with heteroscedastic variance, where it will measure the information gain about the mean prediction.

Maximizing information gain then means to simply maximize the marginal entropy of the ensemble prediction. For this, we make the following observation: the marginal entropy is maximized when the ensemble means are far apart (disagreement) so the modes overlap the least, maximally spreading out probability mass.
As the marginal entropy has no closed-form expression suitable for optimization, we instead use the empirical variance over ensemble means to measure how far apart they are,
\begin{gather}
\begin{aligned}
\operatorname{D}(s_t,a_t) &\triangleq \frac{1}{K-1}\sum_k\big( \mu(w_k,s_t,a_t)-\mu'\big)^2, \\
\mu' &\triangleq \frac{1}{K}\sum_k\mu(w_k,s_t,a_t).
\end{aligned}
\end{gather}%
To summarize, our exploration objective defined in \cref{sec:disagreement}, which maximizes the variance of ensemble means, approximates the information gain and thus should find trajectories that will efficiently reduce the model uncertainty.

\section{Experimental Setup}
\begin{figure*}
    \centering
    \includegraphics[ width=\linewidth]{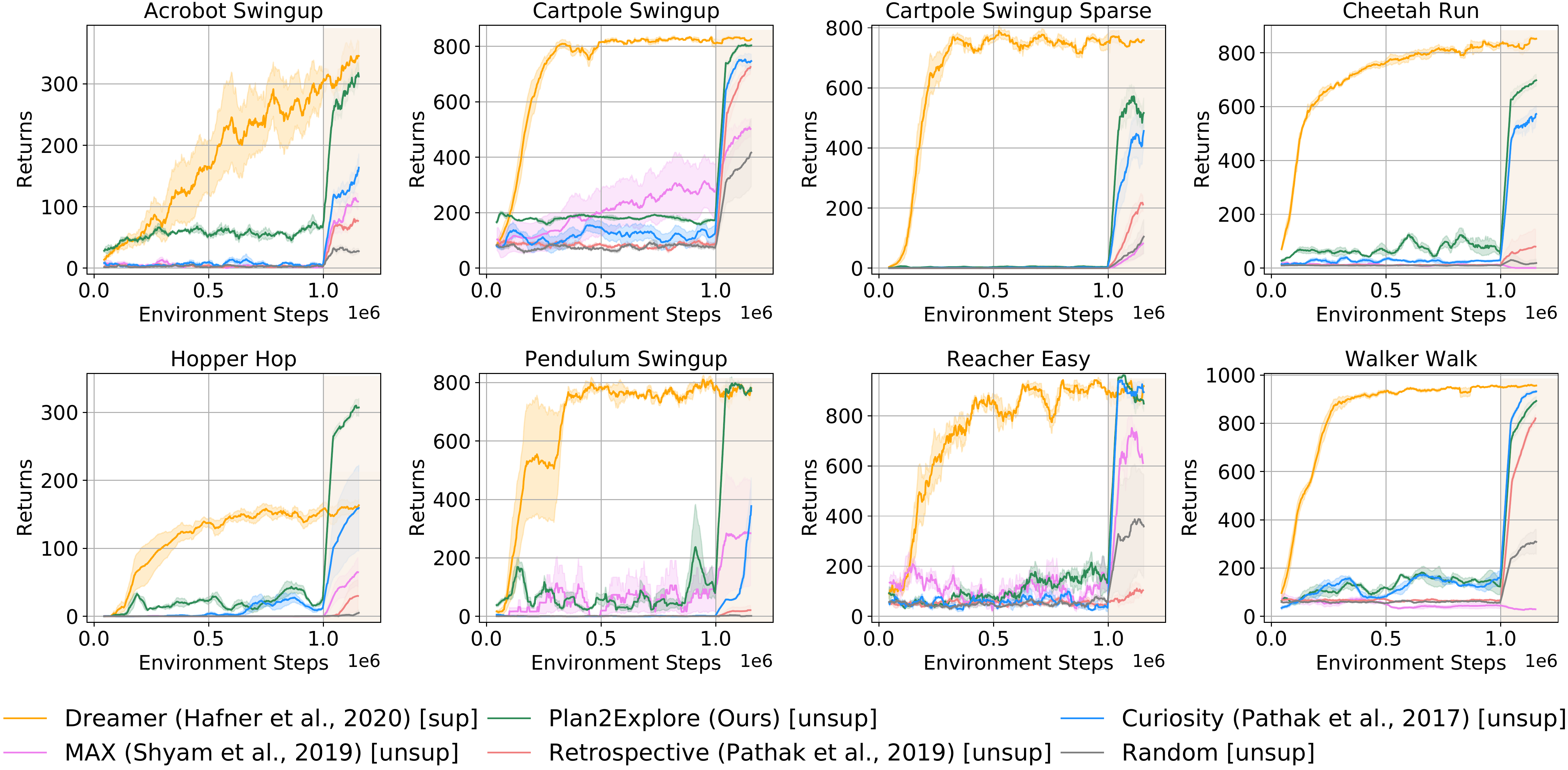}
    \vspace{-0.25in}
    \caption{Performance on few-shot adaptation from raw pixels without state-space input. After the exploration phase of 1M steps (white background), during which the agent does not observe the reward and thus does not solve the task, we let the agent collect a small amount of data from the environment (shaded background). %
    We see that Plan2Explore is able to explore the environment efficiently in only 1000 episodes, and then adapt its behavior immediately after observing the reward. Plan2Explore adapts rapidly, producing effective behavior competitive to state-of-the-art supervised reinforcement learning in just a few collected episodes.}
    \label{fig:adaptation}
\end{figure*}

\paragraph{Environment Details}
We use the DM Control Suite~\cite{deepmindcontrolsuite2018}, a standard benchmark for continuous control. All experiments use visual observations only, of size $64 \times 64 \times 3$ pixels. %
The episode length is 1000 steps and we apply an action repeat of $R=2$ for all the tasks. We run every experiment with three different random seeds with standard deviation shown in the shaded region. Further details are in the appendix.

\vspace{-0.08in}
\paragraph{Implementation}
We use~\cite{hafner2019dreamer} with the original hyperparameters unless specified otherwise to optimize both exploration and task policies of Plan2Explore. We found that additional capacity provided by increasing the hidden size of the GRU in the latent dynamics model to $400$ and the deterministic and stochastic components of the latent space to $60$ helped performance. For a fair comparison, we maintain this model size for Dreamer and other baselines. For latent disagreement, we use an ensemble of $5$ one-step prediction models implemented as $2$ hidden-layer MLP. Full details are in the appendix.

\vspace{-0.08in}
\paragraph{Baselines}
We compare our agent to a state-of-the-art task-oriented agent that receives rewards throughout training, Dreamer ~\cite{hafner2019dreamer}. We also compare to state-of-the-art unsupervised agents: Curiosity~\cite{pathakICMl17curiosity} and Model-based Active Exploration \citep[MAX]{shyam2019max}. Because Curiosity is inefficient during fine-tuning and would not be able to solve a task in a zero-shot way, we adapt it into the model-based setting. We further adapt MAX to work with image observations as~\cite{shyam2019max} only addresses learning from low-dimensional states. We use~\cite{hafner2019dreamer} as the base agent for all methods to provide a fair comparison. We additionally compare to a random data collection policy that uniformly samples from the action space of the environment. %
All methods share the same model hyperparameters to provide a fair comparison. %

\section{Results and Analysis}
Our experiments focus on evaluating whether our proposed Plan2Explore agent efficiently explores and builds a model of the world that allows quick adaptation to solve tasks in zero or few-shot manner. The rest of the subsections are organized in terms of the key scientific questions we would like to investigate as discussed in the introduction.

\subsection{Does the model transfer to solve tasks zero-shot?}
To test whether Plan2Explore has learned a global model of the environment that can be used to solve new tasks, we evaluate the zero-shot performance of our agent. Our agent learns a model without using any task-specific information. After that, a separate downstream agent is trained in imagination, which optimizes the task reward using only the self-supervised world model and no new interaction with the world. To specify the task, we provide the agent with the reward function that is used to label its replay buffer with rewards and train a reward predictor. This process is described in the \cref{alg:adaptation}, with step 10 omitted.

In \cref{fig:zero-shot}, we compare the zero-shot performance of our downstream agent with different amounts of exploration data. This is done by training the downstream agent in imagination at each training checkpoint. The same architecture and hyper-parameters are used for all the methods for a fair comparison. We see that Plan2Explore overall performs better than prior state-of-the-art exploration strategies from high dimensional pixel input, sometimes being the only successful unsupervised method. Moreover, the zero-shot performance of Plan2Explore is competitive to Dreamer, even outperforming it in the hopper hop task.

Plan2Explore was able to successfully learn a good model of the environment and efficiently derive task-oriented behaviors from this model. We emphasize that Plan2Explore explores without task rewards, and Dreamer is the oracle as it is given task rewards during exploration. Yet, Plan2Explore almost matches the performance of this oracle.

\begin{figure*}
    \centering
    \includegraphics[width=\linewidth]{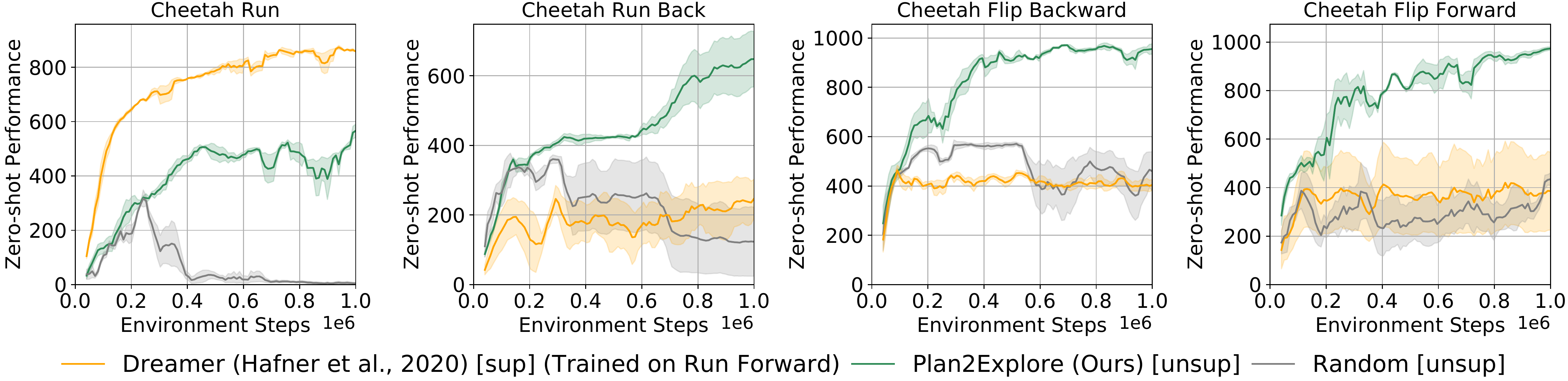}
    \vspace{-0.25in}
    \caption{Do task-specific models generalize? We test Plan2Explore on zero-shot performance on four different tasks in the cheetah environment from raw pixels without state-space input. Throughout the exploration, we take snapshots of policy to plot its zero-shot performance. In addition to random exploration, we compare to an oracle agent, Dreamer, that uses the data collected when trained on the run forward task with rewards. Although Dreamer trained on 'run forward' is able to solve the task it is trained on, it struggles on the other tasks, indicating that it has not learned a global world model.}
    \label{fig:multitask}
\end{figure*}

\subsection{How much task-specific interaction is needed for finetuning to reach the supervised oracle?}
While zero-shot learning might suffice for some tasks, in general, we will want to adapt our model of the world to task-specific information. In this section, we test whether few-shot adaptation of the model to a particular task is competitive to training a fully supervised task-specific model.
To adapt our model, we only add $100-150$ supervised episodes which falls under `few-shot' adaptation.
Furthermore, in this setup, to evaluate the data efficiency of Plan2Explore we set the number of exploratory episodes to only 1000.

In the exploration phase of \cref{fig:adaptation}, i.e., left of the vertical line, our agent does not aim to solve the task, as it is still unknown, however, we expect that during some period of exploration it will \textit{coincidentally} achieve higher rewards as it explores the parts of the state space relevant for the task. The performance of unsupervised methods is coincidental until 1000 episodes and then it switches to task-oriented behavior for the remaining 150 episodes, while for supervised, it is task-oriented throughout. That's why we see a big jump for unsupervised methods where the shaded region begins.

In the few-shot learning setting, Plan2Explore eventually performs competitively to Dreamer on all tasks, significantly outperforming it on the hopper task. Plan2Explore is also able to adapt quickly or similar to other unsupervised agents on all tasks. These results show that a self-supervised agent, when presented with a task specification, should be able to rapidly adapt its model to the task information, matching or outperforming the fully supervised agent trained only for that task. Moreover, Plan2Explore is able to learn this general model with a small number of samples, matching Dreamer, which is fully task-specific, in data efficiency.
This shows the potential of an unsupervised pre-training in reinforcement learning. Please refer to the appendix for detailed quantitative results.

\subsection{Do self-supervised models generalize better than supervised task-specific models?}
If the quality of our learned model is good, it should be transferable to multiple tasks. In this section, we test the quality of the learned model on generalization to multiple tasks in the same environment. %
We devise a set of three new tasks for the Cheetah environment, specifically, running backward, flipping forward, and flipping backward. %
We evaluate the zero-shot performance of Plan2Explore and additionally compare it to a Dreamer agent that is only allowed to collect data on the running forward task and then tested on zero-shot performance on the three other tasks. %

\cref{fig:multitask} shows that while Dreamer performs well on the task it is trained on, running forward, it fails to solve all other tasks, performing comparably to random exploration. It even fails to generalize to the running backward task. In contrast, Plan2Explore performs well across all tasks, outperforming Dreamer on the other three tasks. This indicates that the model learned by Plan2Explore is indeed global, while the model learned by Dreamer, which is task-oriented, fails to generalize to different tasks.

\subsection{What is the advantage of maximizing expected novelty in comparison to retrospective novelty?}

\label{sec:retrospective}

Our Plan2Explore agent is able to measure expected novelty by imagining future states that have not been visited yet. A model-free agent, in contrast, is only trained on the states from the replay buffer, and only gets to see the novelty in retrospect, after the state has been visited. Here, we evaluate the advantages of computing expected versus retrospective novelty by comparing Plan2Explore to a one-step planning agent. The one-step planning agent is not able to plan to visit states that are more than one step away from the replay buffer and is somewhat similar to a Q-learning agent with a particular parametrization of the Q-function. We refer to this approach as Retrospective Disagreement. \cref{fig:zero-shot,fig:adaptation} show the performance of this approach. Our agent achieves superior performance, which is consistent with our intuition about the importance of computing expected novelty.

\section{Related Work}

\paragraph{Exploration}
Exploration is crucial for efficient reinforcement learning~\cite{kakade2002approximately}. In tabular settings, it is efficiently addressed with exploration bonuses based on state visitation counts~\cite{strehl08,jaksch2010near} or fully Bayesian approaches~\cite{duff2002optimal,poupart2006analytic}, however, these approaches are hard to generalize to high-dimensional inputs, such as images. Recently methods are able to scale early ideas to high dimensional data via pseudo-count measures of state visitation~\cite{bellemare2016unifying,ostrovski2017count}. \citet{osband2016deep} derived an efficient approximation to the Thompson sampling procedure via ensembles of Q-functions. \citet{osband2018randomized,lowrey2018plan} use ensembles of Q-functions to track the posterior of the value functions. In contrast to these task-oriented methods, our approach uses neither reward nor state at training time.

\vspace{-1ex}
\paragraph{Self-Supervised RL}
One way to learn skills without extrinsic rewards is to use intrinsic motivation as sole objective~\cite{oudeyer2009intrinsic}.
Practical examples of such approaches focus on maximizing prediction error as curiosity bonus~\cite{pathakICMl17curiosity,burda2018large,haber2018learning}. These approaches can also be understood as maximizing the agent's surprise~\cite{schmidhuber1991curious,josh_surprise}. Similar to our work, other recent approaches use the notion of model disagreement to encourage visiting states with the highest potential to improve the model~\cite{burda2018rnd,pathak2019self}, motivated by the active learning literature~\cite{Seung1992,mccallumzy1998employing}. An alternative is to explore by generating goals from prior experience~\cite{andrychowicz2017hindsight,nair2018imagined}. However, most of these approaches are model-free and expensive to fine-tune to a new task, requiring millions of environment steps for fine-tuning.

\vspace*{-1ex}
\paragraph{Model-based Control}
Early work on model-based reinforcement learning used Gaussian processes and time-varying linear dynamical systems and has shown significant improvements in data efficiency over model-free agents~\cite{kaelbling1996reinforcement,deisenroth2011pilco,levine2013guided} when low-dimensional state information is available.
Recent work on latent dynamics models has shown that model-based agents can achieve performance competitive with model-free agents while attaining much higher data efficiency, and even scale to high-dimensional observations~\cite{chua2018deep,buesing2018learning,ebert2018visual,ha2018world,hafner2018planet,nagabandi2019deep}. We base our agent on a state-of-the-art model-based agent, Dreamer~\cite{hafner2019dreamer}, and use it to perform self-supervised exploration in order to solve tasks in a few-shot manner.

Certain prior work has considered model-based exploration~\cite{amos2018awareness, sharma2019dynamics}, but was not shown to scale to complex visual observations, only using proprioceptive information. Other work~\cite{ebert2018visual,pathak2018zero} has demonstrated the possibility of self-supervised model-based learning with visual observations. However, these approaches do not integrate exploration and model learning together, instead of performing them in stages~\cite{pathak2018zero} or just using random exploration~\cite{ebert2018visual}, which makes them difficult to scale to long-horizon problems.

The idea of actively exploring to collect the most informative data goes back to the formulation of the information gain~\cite{lindley1956expinfogain}. \citet{mackay1992infogain} described how a learning system might optimize Bayesian objectives for active data selection based on the information gain. \citet{sun2011planning} derived a model-based reinforcement learning agent that can optimize the infinite-horizon information gain and experimented with it in tabular settings.
The closest works to ours are \citet{shyam2019max, henaff2019explicit}, which use a measurement of disagreement or information gain through ensembles of neural networks in order to incentivize exploration. However, these approaches are restricted to setups where low-dimensional states are available, whereas we design a latent state approach that scales to high-dimensional observations. Moreover, we provide a theoretical connection between information gain and model disagreement. Concurrent with us,~\citet{ball2020ready} discuss the connection between information gain and model disagreement for task-specific exploration from low-dimensional state space.

\section{Discussion}
We presented Plan2Explore, a self-supervised reinforcement learning method that learns a world model of its environment through unsupervised exploration and uses this model to solve tasks in a zero-shot or few-shot manner. We derived connections of our method to the expected information gain, a principled objective for exploration. Building on recent work on learning dynamics models and behaviors from images, we constructed a model-based zero-shot reinforcement learning agent that was able to achieve state-of-the-art zero-shot task performance on the DeepMind Control Suite. Moreover, the agent's zero-shot performance was competitive to Dreamer, a state-of-the-art supervised reinforcement learning agent on some tasks, with the few-shot performance eventually matching or outperforming the supervised agent. By presenting a method that can learn effective behavior for many different tasks in a scalable and data-efficient manner, we hope this work constitutes a step toward building scalable real-world reinforcement learning systems.
\clearpage
\paragraph{Acknowledgements}
We thank Rowan McAllister, Aviral Kumar, Vijay Balasumbramanian, and the members of GRASP for fruitful discussions. This work was supported in part by NSF-IIS-1703319, ONR N00014-17-1-2093, ARL DCIST CRA W911NF-17-2-0181, Curious Minded Machines grant from Honda Research and DARPA Machine Common Sense grant.

\bibliographystyle{icml2020}
\bibliography{main}

\begin{thebibliography}{55}
\providecommand{\natexlab}[1]{#1}
\providecommand{\url}[1]{\texttt{#1}}
\expandafter\ifx\csname urlstyle\endcsname\relax
  \providecommand{\doi}[1]{doi: #1}\else
  \providecommand{\doi}{doi: \begingroup \urlstyle{rm}\Url}\fi

\bibitem[Achiam \& Sastry(2017)Achiam and Sastry]{josh_surprise}
Achiam, J. and Sastry, S.
\newblock Surprise-based intrinsic motivation for deep reinforcement learning.
\newblock \emph{arXiv:1703.01732}, 2017.

\bibitem[Amos et~al.(2018)Amos, Dinh, Cabi, Rothörl, Muldal, Erez, Tassa,
  de~Freitas, and Denil]{amos2018awareness}
Amos, B., Dinh, L., Cabi, S., Rothörl, T., Muldal, A., Erez, T., Tassa, Y.,
  de~Freitas, N., and Denil, M.
\newblock Learning awareness models.
\newblock In \emph{ICLR}, 2018.

\bibitem[Andrychowicz et~al.(2017)Andrychowicz, Wolski, Ray, Schneider, Fong,
  Welinder, McGrew, Tobin, Abbeel, and Zaremba]{andrychowicz2017hindsight}
Andrychowicz, M., Wolski, F., Ray, A., Schneider, J., Fong, R., Welinder, P.,
  McGrew, B., Tobin, J., Abbeel, P., and Zaremba, W.
\newblock Hindsight experience replay.
\newblock In \emph{NIPS}, 2017.

\bibitem[Ball et~al.(2020)Ball, Parker-Holder, Pacchiano, Choromanski, and
  Roberts]{ball2020ready}
Ball, P., Parker-Holder, J., Pacchiano, A., Choromanski, K., and Roberts, S.
\newblock Ready policy one: World building through active learning.
\newblock \emph{arXiv preprint arXiv:2002.02693}, 2020.

\bibitem[Bellemare et~al.(2016)Bellemare, Srinivasan, Ostrovski, Schaul,
  Saxton, and Munos]{bellemare2016unifying}
Bellemare, M., Srinivasan, S., Ostrovski, G., Schaul, T., Saxton, D., and
  Munos, R.
\newblock Unifying count-based exploration and intrinsic motivation.
\newblock In \emph{NIPS}, 2016.

\bibitem[Breiman(1996)]{breiman1996bagging}
Breiman, L.
\newblock Bagging predictors.
\newblock \emph{Machine learning}, 24\penalty0 (2):\penalty0 123--140, 1996.

\bibitem[Buesing et~al.(2018)Buesing, Weber, Racaniere, Eslami, Rezende,
  Reichert, Viola, Besse, Gregor, Hassabis, et~al.]{buesing2018learning}
Buesing, L., Weber, T., Racaniere, S., Eslami, S., Rezende, D., Reichert,
  D.~P., Viola, F., Besse, F., Gregor, K., Hassabis, D., et~al.
\newblock Learning and querying fast generative models for reinforcement
  learning.
\newblock \emph{arXiv preprint arXiv:1802.03006}, 2018.

\bibitem[Burda et~al.(2018)Burda, Edwards, Storkey, and Klimov]{burda2018rnd}
Burda, Y., Edwards, H., Storkey, A., and Klimov, O.
\newblock Exploration by random network distillation.
\newblock \emph{arXiv preprint arXiv:1810.12894}, 2018.

\bibitem[Burda et~al.(2019)Burda, Edwards, Pathak, Storkey, Darrell, and
  Efros]{burda2018large}
Burda, Y., Edwards, H., Pathak, D., Storkey, A., Darrell, T., and Efros, A.~A.
\newblock Large-scale study of curiosity-driven learning.
\newblock \emph{ICLR}, 2019.

\bibitem[Chua et~al.(2018)Chua, Calandra, McAllister, and Levine]{chua2018deep}
Chua, K., Calandra, R., McAllister, R., and Levine, S.
\newblock Deep reinforcement learning in a handful of trials using
  probabilistic dynamics models.
\newblock \emph{arXiv preprint arXiv:1805.12114}, 2018.

\bibitem[Deisenroth \& Rasmussen(2011)Deisenroth and
  Rasmussen]{deisenroth2011pilco}
Deisenroth, M. and Rasmussen, C.~E.
\newblock Pilco: A model-based and data-efficient approach to policy search.
\newblock In \emph{ICML}, 2011.

\bibitem[Duff \& Barto(2002)Duff and Barto]{duff2002optimal}
Duff, M.~O. and Barto, A.
\newblock \emph{Optimal Learning: Computational procedures for Bayes-adaptive
  Markov decision processes}.
\newblock PhD thesis, University of Massachusetts at Amherst, 2002.

\bibitem[Ebert et~al.(2018)Ebert, Finn, Dasari, Xie, Lee, and
  Levine]{ebert2018visual}
Ebert, F., Finn, C., Dasari, S., Xie, A., Lee, A., and Levine, S.
\newblock Visual foresight: Model-based deep reinforcement learning for
  vision-based robotic control.
\newblock \emph{arXiv:1812.00568}, 2018.

\bibitem[Eysenbach et~al.(2018)Eysenbach, Gupta, Ibarz, and
  Levine]{eysenbach2018diversity}
Eysenbach, B., Gupta, A., Ibarz, J., and Levine, S.
\newblock Diversity is all you need: Learning skills without a reward function.
\newblock \emph{arXiv:1802.06070}, 2018.

\bibitem[Gal(2016)]{gal2016uncertainty}
Gal, Y.
\newblock Uncertainty in deep learning.
\newblock \emph{University of Cambridge}, 1:\penalty0 3, 2016.

\bibitem[Ha \& Schmidhuber(2018)Ha and Schmidhuber]{ha2018world}
Ha, D. and Schmidhuber, J.
\newblock World models.
\newblock \emph{arXiv preprint arXiv:1803.10122}, 2018.

\bibitem[Haber et~al.(2018)Haber, Mrowca, Wang, Fei-Fei, and
  Yamins]{haber2018learning}
Haber, N., Mrowca, D., Wang, S., Fei-Fei, L.~F., and Yamins, D.~L.
\newblock Learning to play with intrinsically-motivated, self-aware agents.
\newblock In \emph{NeurIPS}, 2018.

\bibitem[Hafner et~al.(2019)Hafner, Lillicrap, Fischer, Villegas, Ha, Lee, and
  Davidson]{hafner2018planet}
Hafner, D., Lillicrap, T., Fischer, I., Villegas, R., Ha, D., Lee, H., and
  Davidson, J.
\newblock Learning latent dynamics for planning from pixels.
\newblock \emph{ICML}, 2019.

\bibitem[Hafner et~al.(2020)Hafner, Lillicrap, Ba, and
  Norouzi]{hafner2019dreamer}
Hafner, D., Lillicrap, T., Ba, J., and Norouzi, M.
\newblock Dream to control: Learning behaviors by latent imagination.
\newblock \emph{ICLR}, 2020.

\bibitem[Henaff(2019)]{henaff2019explicit}
Henaff, M.
\newblock Explicit explore-exploit algorithms in continuous state spaces.
\newblock In \emph{NeurIPS}, 2019.

\bibitem[Jaksch et~al.(2010)Jaksch, Ortner, and Auer]{jaksch2010near}
Jaksch, T., Ortner, R., and Auer, P.
\newblock Near-optimal regret bounds for reinforcement learning.
\newblock \emph{JMLR}, 2010.

\bibitem[Kaelbling et~al.(1996)Kaelbling, Littman, and
  Moore]{kaelbling1996reinforcement}
Kaelbling, L.~P., Littman, M.~L., and Moore, A.~W.
\newblock Reinforcement learning: A survey.
\newblock \emph{Journal of artificial intelligence research}, 1996.

\bibitem[Kakade \& Langford(2002)Kakade and Langford]{kakade2002approximately}
Kakade, S. and Langford, J.
\newblock Approximately optimal approximate reinforcement learning.
\newblock In \emph{ICML}, volume~2, pp.\  267--274, 2002.

\bibitem[Kingma \& Welling(2013)Kingma and Welling]{kingma2013vae}
Kingma, D.~P. and Welling, M.
\newblock Auto-encoding variational bayes.
\newblock \emph{arXiv preprint arXiv:1312.6114}, 2013.

\bibitem[Klyubin et~al.(2005)Klyubin, Polani, and
  Nehaniv]{klyubin2005empowerment}
Klyubin, A.~S., Polani, D., and Nehaniv, C.~L.
\newblock Empowerment: A universal agent-centric measure of control.
\newblock In \emph{Evolutionary Computation}, 2005.

\bibitem[Lakshminarayanan et~al.(2017)Lakshminarayanan, Pritzel, and
  Blundell]{lakshminarayanan2017simple}
Lakshminarayanan, B., Pritzel, A., and Blundell, C.
\newblock Simple and scalable predictive uncertainty estimation using deep
  ensembles.
\newblock In \emph{NIPS}, 2017.

\bibitem[Lehman \& Stanley(2011)Lehman and Stanley]{lehman2011evolving}
Lehman, J. and Stanley, K.~O.
\newblock Evolving a diversity of virtual creatures through novelty search and
  local competition.
\newblock In \emph{Proceedings of the 13th annual conference on Genetic and
  evolutionary computation}, 2011.

\bibitem[Levine \& Koltun(2013)Levine and Koltun]{levine2013guided}
Levine, S. and Koltun, V.
\newblock Guided policy search.
\newblock In \emph{International Conference on Machine Learning}, pp.\  1--9,
  2013.

\bibitem[Lindley(1956)]{lindley1956expinfogain}
Lindley, D.~V.
\newblock On a measure of the information provided by an experiment.
\newblock \emph{The Annals of Mathematical Statistics}, pp.\  986--1005, 1956.

\bibitem[Lowrey et~al.(2018)Lowrey, Rajeswaran, Kakade, Todorov, and
  Mordatch]{lowrey2018plan}
Lowrey, K., Rajeswaran, A., Kakade, S., Todorov, E., and Mordatch, I.
\newblock Plan online, learn offline: Efficient learning and exploration via
  model-based control.
\newblock \emph{arXiv preprint arXiv:1811.01848}, 2018.

\bibitem[MacKay(1992)]{mackay1992infogain}
MacKay, D.~J.
\newblock Information-based objective functions for active data selection.
\newblock \emph{Neural computation}, 4\penalty0 (4):\penalty0 590--604, 1992.

\bibitem[McCallumzy \& Nigamy(1998)McCallumzy and
  Nigamy]{mccallumzy1998employing}
McCallumzy, A.~K. and Nigamy, K.
\newblock Employing em and pool-based active learning for text classification.
\newblock In \emph{ICML}, 1998.

\bibitem[Nagabandi et~al.(2019)Nagabandi, Konoglie, Levine, and
  Kumar]{nagabandi2019deep}
Nagabandi, A., Konoglie, K., Levine, S., and Kumar, V.
\newblock Deep dynamics models for learning dexterous manipulation.
\newblock \emph{arXiv preprint arXiv:1909.11652}, 2019.

\bibitem[Nair et~al.(2018)Nair, Pong, Dalal, Bahl, Lin, and
  Levine]{nair2018imagined}
Nair, A.~V., Pong, V., Dalal, M., Bahl, S., Lin, S., and Levine, S.
\newblock Visual reinforcement learning with imagined goals.
\newblock In \emph{NeurIPS}, 2018.

\bibitem[Osband et~al.(2016)Osband, Blundell, Pritzel, and
  Van~Roy]{osband2016deep}
Osband, I., Blundell, C., Pritzel, A., and Van~Roy, B.
\newblock Deep exploration via bootstrapped dqn.
\newblock In \emph{NIPS}, 2016.

\bibitem[Osband et~al.(2018)Osband, Aslanides, and
  Cassirer]{osband2018randomized}
Osband, I., Aslanides, J., and Cassirer, A.
\newblock Randomized prior functions for deep reinforcement learning.
\newblock In \emph{Advances in Neural Information Processing Systems}, pp.\
  8617--8629, 2018.

\bibitem[Ostrovski et~al.(2018)Ostrovski, Bellemare, Oord, and
  Munos]{ostrovski2017count}
Ostrovski, G., Bellemare, M.~G., Oord, A. v.~d., and Munos, R.
\newblock Count-based exploration with neural density models.
\newblock \emph{ICML}, 2018.

\bibitem[Oudeyer \& Kaplan(2009)Oudeyer and Kaplan]{oudeyer2009intrinsic}
Oudeyer, P.-Y. and Kaplan, F.
\newblock What is intrinsic motivation? a typology of computational approaches.
\newblock \emph{Frontiers in neurorobotics}, 2009.

\bibitem[Oudeyer et~al.(2007)Oudeyer, Kaplan, and Hafner]{oudeyer2007intrinsic}
Oudeyer, P.-Y., Kaplan, F., and Hafner, V.~V.
\newblock Intrinsic motivation systems for autonomous mental development.
\newblock \emph{Evolutionary Computation}, 2007.

\bibitem[Pathak et~al.(2017)Pathak, Agrawal, Efros, and
  Darrell]{pathakICMl17curiosity}
Pathak, D., Agrawal, P., Efros, A.~A., and Darrell, T.
\newblock Curiosity-driven exploration by self-supervised prediction.
\newblock In \emph{ICML}, 2017.

\bibitem[Pathak et~al.(2018)Pathak, Mahmoudieh, Luo, Agrawal, Chen, Shentu,
  Shelhamer, Malik, Efros, and Darrell]{pathak2018zero}
Pathak, D., Mahmoudieh, P., Luo, G., Agrawal, P., Chen, D., Shentu, Y.,
  Shelhamer, E., Malik, J., Efros, A.~A., and Darrell, T.
\newblock Zero-shot visual imitation.
\newblock In \emph{ICLR}, 2018.

\bibitem[Pathak et~al.(2019)Pathak, Gandhi, and Gupta]{pathak2019self}
Pathak, D., Gandhi, D., and Gupta, A.
\newblock Self-supervised exploration via disagreement.
\newblock \emph{ICML}, 2019.

\bibitem[Poupart et~al.(2006)Poupart, Vlassis, Hoey, and
  Regan]{poupart2006analytic}
Poupart, P., Vlassis, N., Hoey, J., and Regan, K.
\newblock An analytic solution to discrete bayesian reinforcement learning.
\newblock In \emph{ICML}, 2006.

\bibitem[Rezende et~al.(2014)Rezende, Mohamed, and Wierstra]{rezende2014vae}
Rezende, D.~J., Mohamed, S., and Wierstra, D.
\newblock Stochastic backpropagation and approximate inference in deep
  generative models.
\newblock \emph{arXiv preprint arXiv:1401.4082}, 2014.

\bibitem[Schmidhuber(1991{\natexlab{a}})]{schmidhuber1991curious}
Schmidhuber, J.
\newblock Curious model-building control systems.
\newblock In \emph{Neural Networks, 1991. 1991 IEEE International Joint
  Conference on}, pp.\  1458--1463. IEEE, 1991{\natexlab{a}}.

\bibitem[Schmidhuber(1991{\natexlab{b}})]{schmidhuber1991possibility}
Schmidhuber, J.
\newblock A possibility for implementing curiosity and boredom in
  model-building neural controllers.
\newblock In \emph{From animals to animats: Proceedings of the first
  international conference on simulation of adaptive behavior},
  1991{\natexlab{b}}.

\bibitem[Seung et~al.(1992)Seung, Opper, and Sompolinsky]{Seung1992}
Seung, H., Opper, M., and Sompolinsky, H.
\newblock Query by committee.
\newblock \emph{COLT}, 1992.

\bibitem[Sharma et~al.(2019)Sharma, Gu, Levine, Kumar, and
  Hausman]{sharma2019dynamics}
Sharma, A., Gu, S., Levine, S., Kumar, V., and Hausman, K.
\newblock Dynamics-aware unsupervised discovery of skills.
\newblock \emph{arXiv preprint arXiv:1907.01657}, 2019.

\bibitem[Shyam et~al.(2019)Shyam, Ja\'{s}kowski, and Gomez]{shyam2019max}
Shyam, P., Ja\'{s}kowski, W., and Gomez, F.
\newblock {Model-Based Active Exploration}.
\newblock In \emph{ICML}, 2019.

\bibitem[Strehl \& Littman(2008)Strehl and Littman]{strehl08}
Strehl, A. and Littman, M.
\newblock An analysis of model-based interval estimation for markov decision
  processes.
\newblock \emph{Journal of Computer and System Sciences}, 2008.

\bibitem[Sun et~al.(2011)Sun, Gomez, and Schmidhuber]{sun2011planning}
Sun, Y., Gomez, F., and Schmidhuber, J.
\newblock Planning to be surprised: Optimal bayesian exploration in dynamic
  environments.
\newblock In \emph{AGI}, 2011.

\bibitem[Sutton(1991)]{sutton1991dyna}
Sutton, R.~S.
\newblock Dyna, an integrated architecture for learning, planning, and
  reacting.
\newblock \emph{ACM SIGART Bulletin}, 2\penalty0 (4):\penalty0 160--163, 1991.

\bibitem[Tassa et~al.(2018)Tassa, Doron, Muldal, Erez, Li, de~Las~Casas,
  Budden, Abdolmaleki, Merel, Lefrancq, Lillicrap, and
  Riedmiller]{deepmindcontrolsuite2018}
Tassa, Y., Doron, Y., Muldal, A., Erez, T., Li, Y., de~Las~Casas, D., Budden,
  D., Abdolmaleki, A., Merel, J., Lefrancq, A., Lillicrap, T., and Riedmiller,
  M.
\newblock Deep{Mind} control suite.
\newblock Technical report, DeepMind, January 2018.
\newblock URL \url{https://arxiv.org/abs/1801.00690}.

\bibitem[Watter et~al.(2015)Watter, Springenberg, Boedecker, and
  Riedmiller]{watter2015embed}
Watter, M., Springenberg, J., Boedecker, J., and Riedmiller, M.
\newblock Embed to control: A locally linear latent dynamics model for control
  from raw images.
\newblock In \emph{NIPS}, 2015.

\bibitem[Zhang et~al.(2019)Zhang, Vikram, Smith, Abbeel, Johnson, and
  Levine]{zhang2018solar}
Zhang, M., Vikram, S., Smith, L., Abbeel, P., Johnson, M., and Levine, S.
\newblock Solar: deep structured representations for model-based reinforcement
  learning.
\newblock In \emph{ICML}, 2019.

\end{thebibliography}

\clearpage
\appendix

\section{Appendix}
\begin{figure*}[!ht]
    \centering
    \includegraphics[width=\linewidth]{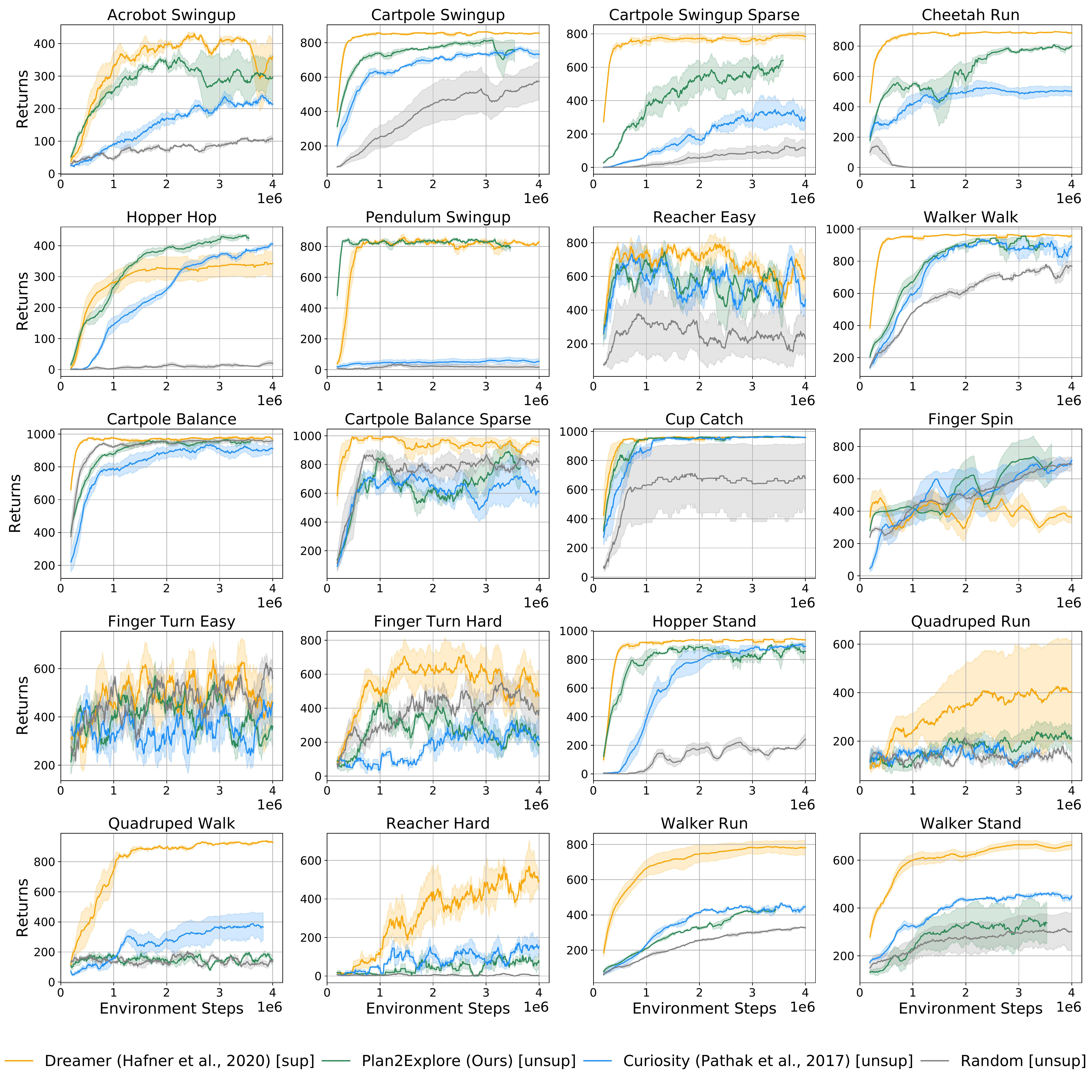}
    \caption{We evaluate the zero-shot performance of the self-supervised agents as well as supervised performance of Dreamer on all tasks from the DM control suite. All agents operate from raw pixels. The experimental protocol is the same as in \cref{fig:zero-shot} of the main paper. To produce this plot, we take snapshots of the agent throughout exploration to train a task policy on the downstream task and plot its zero-shot performance. We use the same hyperparameters for all environments. We see that Plan2Explore achieves state-of-the-art zero-shot task performance on a range of tasks. Moreover, even though Plan2Explore is a self-supervised agent, it demonstrates competitive performance to Dreamer~\cite{hafner2019dreamer}, a state-of-the-art supervised reinforcement learning agent. This shows that self-supervised exploration is competitive to task-specific approaches in these continuous control tasks.}
    \label{fig:zeroshot_supp}
\end{figure*}

\paragraph{Results DM Control Suite}
In \cref{fig:zeroshot_supp}, we show the performance of our agent on all 20 DM Control Suite tasks from pixels. In addition, we show videos corresponding to all the plots on the project website: \url{https://ramanans1.github.io/plan2explore/}

\paragraph{Convention for plots}
We run every experiment with three different random seeds. The shaded area of the graphs shows the standard deviation in performance. All plot curves are smoothed with a moving mean that takes into account a window of the past 20 data points. Only \cref{fig:multitask} was smoothed with a window of past 5 data points so as to provide cleaner looking plots that indicate the general trend. Low variance in all the curves consistently across all figures suggests that our approach is very reproducible.

\paragraph{Rewards of new tasks}
To test the generalization performance of the our agent, we define three new tasks in the Cheetah environment:
\vspace*{-1ex}
\begin{itemize}
\item\textbf{Cheetah Run Backward}\quad
Analogous to the forward running task, the reward $r$ is linearly proportional to the backward velocity $v_b$ up to a maximum of $10 \mathrm{m/s}$, which means $r(v_b) = \max(0,\min(v_b/10,1))$, where $v_b=-v$ and $v$ is the forward velocity of the Cheetah.
\item\textbf{Cheetah Flip Backward}\quad
The reward $r$ is linearly proportional to the backward angular velocity $\omega_b$ up to a maximum of $5 \mathrm{rad/s}$, which means $r(\omega_b) = \max(0,\min(\omega_b/5,1))$, where $\omega_b = -\omega$ and $\omega$ is the angular velocity about the positive Z-axis, as defined in DeepMind Control Suite.
\item\textbf{Cheetah Flip Forward}\quad
The reward $r$ is linearly proportional to the forward angular velocity $\omega$ up to a maximum of $5 \mathrm{rad/s}$, which means $r(\omega) = \max(0,\min(\omega/5,1))$.
\end{itemize}

\paragraph{Environment}
We use the DeepMind Control Suite~\cite{deepmindcontrolsuite2018} tasks, a standard benchmark of tasks for continuous control agents. All experiments are performed with only visual observations. We use RGB visual observations with $64 \times 64$ resolution. We have selected a diverse set of 8 tasks that feature sparse rewards, high dimensional action spaces, and environments with unstable equilibria and environments that require a long planning horizon. We use episode length of 1000 steps and a fixed action repeat of $R=2$ for all the tasks.

\paragraph{Agent implementation}
For implementing latent disagreement, we use an ensemble of $5$ one-step prediction models with a $2$ hidden-layer MLP, which takes in the RNN-state of RSSM and the action as inputs, and predicts the encoder features, which have a dimension of $1024$. We scale the disagreement of the predictions by $10,000$ for the final intrinsic reward, this was found to increase performance in some environments. We do not normalize the rewards, both extrinsic and intrinsic. This setup for the one-step model was chosen over $3$ other variants, in which we tried predicting the deterministic, stochastic, and the combined features of RSSM respectively. The performance benefits of this ensemble over the variants potentially come from the large parametrization that comes with predicting the large encoder features.

\paragraph{Baselines}
We note that while Curiosity~\cite{pathakICMl17curiosity} uses $L_2$ loss to train the model, the RSSM loss is different (see~\cite{hafner2018planet}); we use the full RSSM loss as the intrinsic reward for the Curiosity comparison, as we found it produces the best performance. Note that this reward can only be computed when ground truth data is available and needs a separate reward predictor to optimize it in a model-based fashion. %

\begin{table*}
\caption{Zero-shot performance at 3.5 million environment steps (corresponding to 1.75 agent steps times 2 for action repeat).  We report the average performance of the last 20 episodes before the 3.5 million steps point. The performance is computed by executing the mode of the actor without action noise. Among the agents that receive no task rewards, the highest performance of each task is highlighted. The corresponding training curves are visualized in \cref{fig:zeroshot_supp}.}
\begin{tabularx}{\textwidth}{
@{}
X
>{\raggedleft\arraybackslash}p{5em}
>{\raggedleft\arraybackslash}p{5em}
>{\raggedleft\arraybackslash}p{5em}
>{\raggedleft\arraybackslash}p{5em}
>{\raggedleft\arraybackslash}p{5em}
>{\raggedleft\arraybackslash}p{5em}
@{}}
\toprule
\textbf{Zero-shot performance} & \textbf{Plan2Explore}  & \textbf{Curiosity}  & \textbf{Random} & \textbf{MAX} & \textbf{Retrospective} & \textbf{Dreamer}  \\
\midrule
Task-agnostic experience & 3.5M & 3.5M & 3.5M & 3.5M & 3.5M & $-$ \\
Task-specific experience & $-$ & $-$ & $-$ & $-$ & $-$ & 3.5M \\
\midrule
    Acrobot Swingup         &           \textbf{280.23}  &           219.55  & 107.38 & 64.30   &   110.84 &   408.27 \\
    Cartpole Balance         &  950.97 &         917.10  & \textbf{963.40} & $-$  &   $-$  &   970.28 \\
    Cartpole Balance  Sparse& \textbf{860.38} &         695.83  & 764.48 & $-$     & $-$   &   926.9 \\
    Cartpole Swingup        &  \textbf{759.65} &         747.488  & 516.04 & 144.05   &   700.59 &   855.55 \\
    Cartpole Swingup Sparse  &          \textbf{602.71}  &  324.5  & 94.89 & 9.23   & 180.85 &   789.79 \\
    Cheetah Run             &          \textbf{784.45}  &         495.55  & 0.78 & 0.76   &   9.11 &   888.84 \\
    Cup Catch               &  \textbf{962.81} &         \textbf{963.13}  & 660.35 & $-$  &   $-$  &   963.4 \\
    Finger Spin            &  655.4 &         661.96  &         \textbf{676.5} & $-$  &  $-$    &  333.73 \\
    Finger Turn Easy   &  401.64 &         266.96  &         \textbf{495.21}  & $-$      &  $-$  &   551.31 \\
    Finger Turn Hard       &  270.83 &         289.65  &         \textbf{464.01}  & $-$  &  $-$   &  435.56\\
    Hopper Hop               &          \textbf{432.58}  &           389.64  & 12.11 & 17.39  &    41.32 &   336.57 \\
    Hopper Stand             &  841.53 &           \textbf{889.87}  & 180.86 & $-$  &    $-$ &   923.74 \\
    Pendulum Swingup         &          \textbf{792.71}  &           56.80  & 16.96 & 748.53  &    1.383 &   829.21 \\
    Quadruped Run           &             \textbf{223.96}  &         164.02  & 139.53 & $-$  &      $-$  &   373.25 \\
    Quadruped Walk          &             182.87  &         \textbf{368.45}  & 129.73 & $-$  &      $-$  &   921.25 \\
    Reacher Easy             &  \textbf{530.56} &         416.31  & 229.23 & 242.13  &    230.68 &   544.15 \\
    Reacher Hard             &  66.76  &          \textbf{123.5}  &         4.10  & $-$  &   $-$  &   438.34 \\
    Walker Run              &          429.30  &         \textbf{446.45}  & 318.61 & $-$  &   $-$  &   783.95 \\
    Walker Stand             &  331.20 &         \textbf{459.29}  & 301.65 & $-$  &   $-$  &   655.80 \\
    Walker Walk             &  \textbf{911.04} & 889.17 & 766.41 & 148.02   &   538.84 &   965.51 \\
\midrule
    Task Average            &          \textbf{563.58}  &  489.26 & 342.11 & $-$      & $-$    &  694.77 \\
\bottomrule
\end{tabularx}

\end{table*}

\begin{table*}
\caption{Adaptation performance after 1M task-agnostic environment steps, followed by 150K task-specific environment steps (agent steps are half as much due to the action repeat of 2).  We report the average performance of the last 20 episodes before the 1.15M steps point. The performance is computed by executing the mode of the actor without action noise. Among the self-supervised agents, the highest performance of each task is highlighted. The corresponding training curves are visualized in \cref{fig:adaptation}.}
\begin{tabularx}{\textwidth}{
@{}
X
>{\raggedleft\arraybackslash}p{5em}
>{\raggedleft\arraybackslash}p{5em}
>{\raggedleft\arraybackslash}p{5em}
>{\raggedleft\arraybackslash}p{5em}
>{\raggedleft\arraybackslash}p{5em}
>{\raggedleft\arraybackslash}p{5em}
@{}}
\toprule
\textbf{Adaptation performance} & \textbf{Plan2Explore}  & \textbf{Curiosity}  & \textbf{Random} & \textbf{MAX} & \textbf{Retrospective} & \textbf{Dreamer}  \\
\midrule
Task-agnostic experience & 1M & 1M & 1M & 1M & 1M & $-$ \\
Task-specific experience & 150K & 150K & 150K & 150K & 150K & 1.15M \\
\midrule
    Acrobot Swingup         &           \textbf{312.03}  &           163.71  & 27.54 & 108.39   &   76.92 &   345.51 \\
    Cartpole Swingup        &  \textbf{803.53} &        747.10   & 416.82  & 501.93   &  725.81  &   826.07 \\
    Cartpole Swingup Sparse  &          \textbf{516.56}  &  456.8  & 104.88  & 82.06   & 211.81 &  758.45 \\
    Cheetah Run             &          \textbf{697.80}  &         572.67  & 18.91 & 0.76   &   79.90 &   852.03 \\
    Hopper Hop               &          \textbf{307.16}  &           159.45  & 5.21 & 64.95  &    29.97 &  163.32  \\
    Pendulum Swingup         &          \textbf{771.51}  &           377.51  & 1.45 & 284.53  &  21.23   &   781.36 \\
    Reacher Easy             &  848.65 &         \textbf{894.29}  & 358.56 & 611.65  &  104.03   &   918.86 \\
    Walker Walk             &  892.63 & \textbf{932.03} & 308.51 & 29.39   &  820.54  &   956.53 \\
\midrule
    Task Average            &          \textbf{643.73}  & 537.95  & 155.23  & 210.46      & 258.78    &  700.27 \\
\bottomrule
\end{tabularx}
\end{table*}

\end{document}